\documentclass{llncs}
\usepackage{graphicx} 
\usepackage{booktabs} 
\usepackage{multirow} 
\usepackage{soul} 
\usepackage{xcolor} 
\usepackage[normalem]{ulem} 
\usepackage{mathtools} 

\usepackage{amsmath}
\usepackage{algpseudocode}
\usepackage{algorithm}
\usepackage{placeins}

\title{Universal Vector Neural Machine Translation With Effective Attention}

\author{
Satish Mylapore\inst{1} \and
Ryan Quincy Paul\inst{1} \and
Joshua Yi\inst{1} \and
Robert D. Slater\inst{1}
}

\institute{
Master of Science in Data Science, Southern Methodist University,
Dallas TX 75275 USA 
\email{\{smylaporesaravanabhava, rqpaul, jyi, rslater\}@smu.edu}
}

\begin{document}

\maketitle              

\setcounter{footnote}{0}
\begin{abstract}
    Neural Machine Translation (NMT) leverages one or more trained neural networks for the translation of phrases. Sutskever introduced a sequence to sequence based encoder-decoder model which became the standard for NMT based systems. Attention mechanisms were later introduced to address the issues with the translation of long sentences and improving overall accuracy. In this paper, we propose a singular model for Neural Machine Translation based on encoder-decoder models. Most translation models are trained as one model for one translation. We introduce a neutral/universal model representation that can be used to predict more than one language depending on the source and a provided target. Secondly, we introduce an attention model by adding an overall learning vector to the multiplicative model. With these two changes, by using the novel universal model the number of models needed for multiple language translation applications are reduced.

\end{abstract}


\section{Introduction} \label{section_introduction}

Neural Machine Translation (NMT) \cite{kalchbrenner_recurrent_continuous_2013} is a significant recent development in large scale translation \cite{jean_very_large_target_vocab,luong_addressing_the_rare_word_problem_2014}. The traditional translation model introduced by Koehn et al. 2003 \cite{koehn_statistical_phrase_based_translation_2003} was trained on a single large neural model with layers that are trained separately requiring many resources and effort. Today, most industry players have adopted a neural network based machine translation system derived from the Recurrent Neural Network (RNN) encoder-decoder model introduced by Cho et al. 2014 \cite{Cho_2014_learning_phrase_representations_rnn}. For machine translation, the encoder is used with the source language to encode the sentence input into a vector representation for the decoder. The decoder uses the encoded sequence to begin predicting the target sequence. There were several advancements to this model by the introduction of different types of RNNs such as LSTM (Long Short Term Memory) \cite{hochreiter_long_short_term_1997,rumelhart_learning_representations_back_probagating_1988,schwenk_cslm_space_language_toolkit_2013,sundermeyer_lstm_nn_language_2012}, GRU (Gated Recurrent Unit) \cite{Cho_2014_properties_of_nmt}, and Bi-RNN (Bidirectional RNN) \cite{schuster_bidirectional_rnn_1997} which was introduced to address the vanishing gradient problem \cite{pascanu_on_the_difficulty_of_training_2013} that was encountered during the training of the simple recurrent neural network.  

Gated recurrent networks failed to fully resolve the problem of the encoder-decoder network \cite{pouget_abadie_overcoming_the_curse_2014} which is the ability to learn and maintain information of the encoder for longer sentences. This is where the attention mechanism was introduced by Graves et al. 2014 that is based on the cosine similarity of the sentences \cite{graves_neural_turing_2014}, Bahdanau et al. 2014 which concatenates the encoder and decoder information \cite{bahdanau_2014_nmt_align}, and Loung et al. 2015 that uses the dot product of the the encoder and decoder information to score the attention on the target sequence \cite{luong_addressing_the_rare_word_problem_2014,luong_effective_approaches_attention}. The introduction of attention mechanisms increased the scalability of machine translation at the cost of performance during training.

The latest development in the machine translation space is the introduction of the Transformer model by Vaswani et al. 2017
\cite{Vaswani_2017_attention_is_all_you_need}. The Transformer model focuses more on self-attention and fully leverages recurrent networks. It promotes self-attention in both the encoder and the decoder where the encoding of the source sequences are done in parallel. This reduces the training time significantly. The decoder prediction is auto-regressive which means it predicts each word at a time in a regressive state. Vaswani claims that the results of the transformer model has a significant improvement in prediction accuracy when compared to other recent models in the NMT space with the use of a German translation task \cite{Vaswani_2017_attention_is_all_you_need}.

The transformer model is still in the incubation and adoption stage in current industry practice. This is due to its restricted context length during translation (fixed-length context). Furthermore, at present all RNN encoder-decoder based machine translation models still use a single model architecture for a translation job. For example, if a task requires translation from Spanish to English, one model will be trained. Another model would be trained to translate from English to Spanish. One model corresponds to one translation task, hence separate models are required. In this research, we seek to build a singular model to translate multiple languages. For the purpose of this research we have considered English-Spanish and Spanish-English translation using the same model.

All machine translation mechanisms to date use language specific encoders for each source language \cite{hermann_multilingual_distributed_2014}. This paper will detail a novel method of hosting multiple neural machine translation tasks within the same model as follows. Section~\ref{section_related_work} will cover related works on the fundamental concepts of the sequence to sequence Recurrent Neural Network based Encoder-Decoder model, the additive attention model by Bahdanau, and wrap-up with the Dual Learning method introduced by Microsoft. Section~\ref{section_model_architecture} outlines the architecture for the universal vector model and discusses each layer. Section~\ref{section_model_training} discusses the training method for the universal model, while Section~\ref{section_dataset} explains the dataset and how it is used for training. Section~\ref{section_bleu_score} is an overview of the BLEU score. The translation results of the Universal Vector is explained in Section~\ref{section_results} then Section~\ref{section_model_analysis} presents the analysis of the BLEU score, loss results, and attention model performance. Section~\ref{section_limitations_and_future_expansion} goes over limitations and potential steps to take in the future with Section~\ref{section_previously_considered_experiments} discussing previously considered experiments. Finally, the paper is concluded with some parting thoughts on the development of this novel model in Section~\ref{section_conclusion}.

\section{Related Work} \label{section_related_work}

This section will go over the associated work related to building the Universal Vector Neural Machine Translation model. First Recurrent Neural Network Based Encoder-Decoder Models proposed by Sutskever et al. and Cho et al. will be discussed. Next, the attention mechanism first proposed by Bahdanau et al. will be detailed. Finally, the Dual Learning model training approach is explained.

\subsection{Recurrent Neural Network Based Encoder-Decoder Models}
\label{subsection:recurrent_neural_network_based_enc_dec_models}

Many NMTs are built upon the fundamental Recurrent Neural Network (RNN) based Encoder-Decoder model as proposed by Sutskever et al. (2014) and by Cho et al. (2014) \cite{Cho_2014_learning_phrase_representations_rnn,Cho_2014_properties_of_nmt,forcada_recursive_hetero_1997,sutskever_sequence_to_sequence}. This model uses two networks, an encoder and a decoder, to learn sequences of information and make predictions. In this model a sequence of input $\mathrm{x}$ is provided to the encoder, an RNN. An RNN allows for outputs of iterations through a network to be passed on as input to future iterations \cite{gers_learning_to_forget_lstm_1999,lipton_critical_review_rnn_2015,salehinejad_recent_advances_rnn_2018,sherstinsky_fundamentals_rnn_2018}. $\mathrm{x}$ is processed word by word ($1, 2,\dots, t$) over multiple iterations. Each iteration calculates a hidden state that is based on the current word in a phrase ($x_t$) and the hidden states of previous iterations ($h_{t-1}$). This is represented at a high level in Equation~\ref{eq:hidden_states_base_encoder} below with a non-linear equation $q$ calculating hidden states at each position \cite{bahdanau_2014_nmt_align}.

\begin{equation}
  \label{eq:hidden_states_base_encoder}
  h_t = f(x_t, h_{t-1})
\end{equation}

Once all hidden states have been calculated, a function $g$ will then return a single fixed length context vector $c$ with each hidden state as inputs like in Equation~\ref{eq:context_vector_base}  below. $c$ represents the full summary of the output of the encoder network \cite{bahdanau_2014_nmt_align}.

\begin{equation}
  \label{eq:context_vector_base}
  c=q(\{h_i,\dots,h_{T_x}\})
\end{equation}

The output of the encoder, $c$, is then fed into the decoder which is another trained RNN. The decoder emits the prediction for each input $y$ at iteration $t$ where these conditional outputs come together as a probability distribution like below in Equation~\ref{eq:decoder_prediction_output} \cite{bahdanau_2014_nmt_align}.

\begin{equation}
    \label{eq:decoder_prediction_output}
    p(y_t | \{y_1,\dots,y{t-1}\},\mathrm{c}) = g(y_{t-1}, s_t, c)
\end{equation}

$g$ is another non-linear function that takes in the previously predicted words ($y_{t-1}$), the hidden state of the current iteration of the network ($s_t$), and the context vector from above ($c$). $p$ represents a predicted target sequence of words for a given input sequence of words with conditional probability \cite{pascanu_how_to_construct_deep_2014}. This is the basis of the Encoder-Decoder model that has been used heavily in neural machine translation.

\subsection{Attention Mechanism}

Attention mechanisms have gained visibility recently as they are able to improve the performance of translation by helping the encoder and decoder to align by providing guidance on what parts of a large sentence will be most useful in predicting the next word \cite{bahdanau_2014_nmt_align,luong_effective_approaches_attention,Vaswani_2017_attention_is_all_you_need,wu_neural_machine_translation_2016}. In recent years many attention models have been introduced such as Bahdanau et al \cite{bahdanau_2014_nmt_align} which concatenates (referred to as "concat" in Luong, et al., 2015 \cite{luong_effective_approaches_attention} and as "additive attention" in Vaswani, et al., 2017 \cite{Vaswani_2017_attention_is_all_you_need}) forward and backward information from the source. This model changes the fundamental RNN Encoder-Decoder described above in a variety of ways.

The encoder is built using a bi-directional recurrent neural network that contains two models. Each model will compute hidden states in either direction from a given input $x_i$. This will yield two hidden states, $\overrightarrow{h_i}$ and $\overleftarrow{h_i}$. These two hidden states are concatenated together to form a vector $h_i$ as below in Equation~\ref{eq:concatenated_hidden_state_encoder_bahdanau} that will represent the whole sentence emanating out from a given input word and are referred to as annotations \cite{bahdanau_2014_nmt_align}. 

\begin{equation}
    \label{eq:concatenated_hidden_state_encoder_bahdanau}
        h_i = 
        \begin{bmatrix}
          \overrightarrow{h_i} \\
          \overleftarrow{h_i} 
        \end{bmatrix}
\end{equation}

Due to RNNs tendency toward recency bias, the words immediately surrounding around a given input ($x_i$) will be better represented in the input word's annotation ($h_i$). This will be reflected when calculating attention which begins with a replacement to the fixed length context vector $c$ mentioned in Section~\ref{subsection:recurrent_neural_network_based_enc_dec_models}. A new context vector $c_j$ is calculated for every output word $y_j$. This begins with a scoring function $e_{ij}$ which will represent the importance of the hidden state output from the previous iteration of the decoder $s_{j-1}$ to a given annotation $h_i$ represented by Equation~\ref{eq:associated_energy_bahdanau} below \cite{bahdanau_2014_nmt_align}. A higher score will represent higher importance.

\begin{equation}
    \label{eq:associated_energy_bahdanau}
    e_{ij} = a(s_{j-1}, h_i)
\end{equation}

$e_{ij}$ is then fed into a softmax function $\alpha_{ij}$ found below in Equation~\ref{eq:weight_of_each_annotation} which will return a vector of numbers that all sum up to one that represents the weight of each annotation with respect to the given position of $y_j$ \cite{bahdanau_2014_nmt_align}.

\begin{equation}
    \label{eq:weight_of_each_annotation}
    \alpha_{ij} = \frac{\exp(e_{ij})} {{\sum_{k=1}^{T_x}} \exp(e_{jk})}
\end{equation}

Finally, the context vector $c_j$ unique to each word output by the decoder $y_j$ is calculated with the summation found in Equation~\ref{eq:context_vector_bahdanau} below.

\begin{equation}
    \label{eq:context_vector_bahdanau}
    c_j = \sum\limits_{i=1}^{T_x} \alpha_{ij}h_i
\end{equation}

Vector $c_j$ will be used in the calculation of hidden states in the decoder $s_j$ found in Equation~\ref{eq:decoder_hidden_state_bahdanau}. $s_j$, the previously predicted words $y_{j-1}$, and the $c_j$ will then be used in calculating the output of each iteration of the decoder at step $j$ as in Equation~\ref{eq:decoder_with_attention_bahdanau} below. The output is a vector of probabilities of each possible word that could be predicted at $y_j$. The context vector $c_i$ will weigh in words at input position $i$ that scored a higher importance from $e_{ij}$ in Equation~\ref{eq:associated_energy_bahdanau} more than others which represents attention. This in in contrast to taking the whole vector of input words into account at every $j$th position of $y$ \cite{bahdanau_2014_nmt_align}.

\begin{equation}
    \label{eq:decoder_hidden_state_bahdanau}
    s_j = f(s_{j-1}, y_{j-1}, c_j)
\end{equation}

\begin{equation}
    \label{eq:decoder_with_attention_bahdanau}
    p(y_j | \{y_1,\dots,y{j-1}\},x) = g(y_{j-1}, s_j, c_j)
\end{equation}

This is an early implementation of attention proposed by Bahdanau et al. 2014 \cite{bahdanau_2014_nmt_align}. Many other forms of attention have been proposed since. Luong et al refers to Bahdanau's attention mechanism as "global attention." In turn, Luong et al. proposed a "local attention method" that focuses on smaller portions of context instead of applying attention weights on the entire source text \cite{luong_effective_approaches_attention}. The new attention mechanism proposed in this paper combines the two. 

\subsection{Dual Learning}

In a paper proposed by Microsoft Research \cite{xia_2016_dual_learning}, the team considered a dual learning mechanism to handle the complexities in the training data labeling. The dual learning mechanism considers two agents, one agent for the forward translation model (source to target language) and the second agent is considered for the dual translation (target to source language). These models use two different corpora for training which are not parallel data sets. This enables reinforcement learning for the convergence of source and target language. The inputs considered on the Microsoft Research paper are "Monolingual corpora $D_A$ and $D_B$, initial translation models $\theta A B$ and $\theta B A$, language models LMA and LMB, hyper-parameter $\alpha$, beam search size $K$, learning rates $\gamma_{1,t}$, $\gamma_{2,t}$." \cite{xia_2016_dual_learning} The experiment used to test dual training uses two separate models, one for each translation direction.

In this paper, there are two contributions based on RNN Encoder-Decoder based machine translations. First, a neutral/universal vector representation for machine translation is introduced. Then a modified attention mechanism based on global attention mechanism proposed in Luong et al. \cite{luong_effective_approaches_attention} and Bahdanau et al. \cite{bahdanau_2014_nmt_align} is discussed. Finally, testing of the proposed neutral vector representation with modified attention mechanism are examined and the results are presented.

\section{Model Architecture} \label{section_model_architecture}
The architecture of the model is built on top of the basic sequence to sequence model and modified to translate more than one language. A high level architecture diagram is found in Fig.~\ref{model_architecture_1} below. This model contains two networks, an encoder and a decoder, with embedded inputs and outputs for each. It also contains the modified attention mechanism and a Fully Connected Layer. In the current structure, the source text is inserted into the Input Embedding layer which contains the Encoder RNN. There are multiple Input Embedding Layers to handle different source texts such as Spanish, English, German, etc. From the Input Embedding layer, the results (context vectors) are fed into the Target Embedding layer which contains the Decoder RNN along with the modified Attention layer. As is the case with the Encoder portion of the system, there are multiple Target Embedding layers for multiple target languages. Lastly, the output from the Target Embedding layer is passed into the Target Fully Connected layer. The result is a vector of probabilities for words in the target language. From this vector, the predicted phrase is converted from a numeric vector representation to words in a natural language. 

\begin{figure}
    \centering
    \includegraphics[width=10cm]{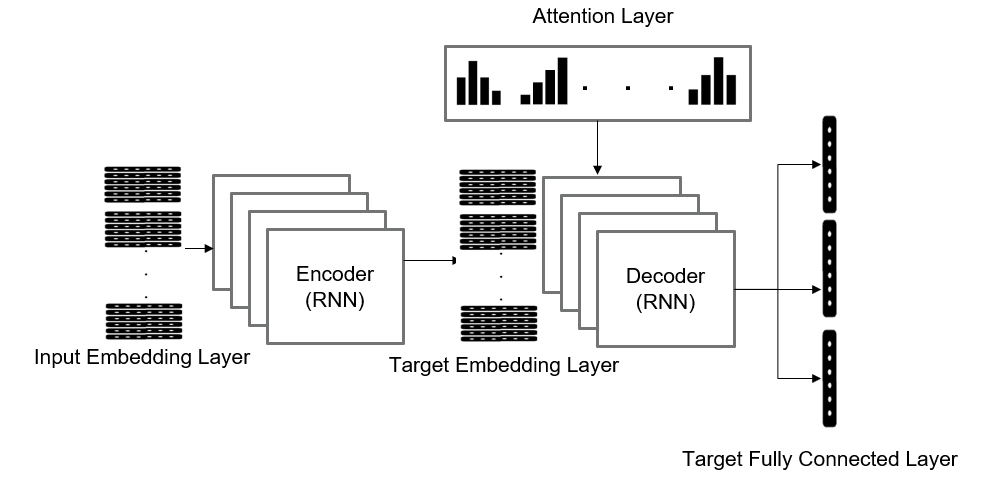}
    \caption{Model architecture detailing the encoder and decoder networks and their inputs.} 
    \label{model_architecture_1}
\end{figure}
\FloatBarrier

\subsection{Embedding Layers}
The model starts with an encoder layer to generate a vector that will be fed to the decoder to generate predictions in a target language. Embedding vectors for the encoder will be built as layers, which are considered to be the source input to the encoding layer as Equation~\ref{eq:architecture_source_embedding_layers} below.

\begin{equation}
    \label{eq:architecture_source_embedding_layers}
    es \sim  es_1, es_2, \dots, es_n
\end{equation}

Here, $es$ is the embedding vector and each number represented by $_s$ represents a different language used as a source for translation. This will be used as the first layer in the encoder network. Similarly, an embedding layer for the decoder is also built as in Equation~\ref{eq:architecture_target_embedding_layers} below.

\begin{equation}
    \label{eq:architecture_target_embedding_layers}
    et \sim  et_1, et_2, \dots, et_n
\end{equation}

$et$ is the embedding vector and each number represented by $_t$ represents a different language used as the target prediction. 

\subsection{Attention Layer}

The modified attention mechanism considers a context vector $\mathrm{c}$ created by the encoder. This vector $\mathrm{c}$ is created based on all the hidden vectors of the hidden states during the encoding phase. This vector has a representation of each word from the source. The attention score used to predict each target word is calculated by the dot product of the hidden value of each prediction and the encoded output \cite{Vaswani_2017_attention_is_all_you_need}. This scoring mechanism is based on the global attention method proposed by Luong et al. \cite{luong_effective_approaches_attention}. Learning weights were introduced into the dot product score, which is calculated using Equation~\ref{eq:global_attention_score_with_learning_weights} below. The purpose of this is to learn the overall weights of the dot product score. 

\begin{equation}
    \label{eq:global_attention_score_with_learning_weights}
    score(h_t, \overline{h}_s) = v_{a}^{\top}(h_t^{\top}W_a \overline{h}_s)
\end{equation}

The context vector is computed by taking the dot product of the encoder output. This is done to add global alignment to the context vector, which will be used to estimate the score of the next prediction.

The attention mechanism will be used to align decoder predictions of the target vector. Attention weights for each target language is defined as Equation~\ref{eq:attention_each_target_language} below. Where $a$ is the attention weight and $t$ is the target language.

\begin{equation}
    \label{eq:attention_each_target_language}
    a_t \sim (a_{t1}, a_{t2}, \dots, a_{tn})
\end{equation}

\subsection{Fully Connected Layer}
The last layer is a fully connected layer, which will be attached to the size of the target language as seen in Equation~\ref{eq:connected_layer_each_target_language} below where $fc$ is a connected layer and $t$ is each target language. The purpose of the fully connected layer is to act as a classifier for each targeted translated text.

\begin{equation}
    \label{eq:connected_layer_each_target_language}
    fc \sim (fc_{t1}, fc_{t2}, \dots, fc_{tn})
\end{equation}

\section{Model Training} \label{section_model_training}

The model training process considers training for each set of translations in a sequence. For this experiment, Spanish and English languages are considered for training with Gated Recurrent Units (GRUs) as the recurrent unit were considered to address the long term dependencies \cite{hochreiter_long_short_term_1997,bengio_learning_long_term_1994,hochreiter_long_term_1991,hochreiter_gradient_flow_2001} where $W1_{en}$, $V_{en}$, $W1_{sp}$, and $V_{sp}$ all act as attention weight matrices, $e_1$ is the embedding layer for Spanish and $e_2$ is the embedding layer for English, $d_1$ is the fully connected layer for Spanish and $d_2$ is the fully connected layer for English. The weight matrices for each gate in the GRU are represented as $\Gamma_{ue}$ and $\Gamma_{ud}$ for the update gates, $\Gamma_{re}$ and $\Gamma_{rd}$ for relevance gates, $c_e$ and $c_d$ for the context gates, and $h_{(te)}$ and $h_{(td)}$ represent the hidden vectors. The weights will be initialized using the Glorot Uniform Initializer \cite{glorot_2010}. The Adam optimization algorithm has been used with a learning rate $\alpha$ and a decayed learning rate $\gamma$. Loss will be measured using the discrete classification methodology which leverages the sparse Softmax cross-entropy with logits loss. Spanish-English will be used as the parallel database where each example of language will be trained in parallel.

A pseudo algorithm of the training process is given below.

\begin{algorithm} \caption{Model Training Process}
    \begin{algorithmic}[1]
    \Require Parallel Dataset $D$ with phrases in both $Lang_A$ and $Lang_B$
    \Repeat
    \ForAll {Phrases $p$ made up of $Lang_A, Lang_B$ in $D$}
    \State \textbf{Block 1:} Encode $Lang_A$ example and compute encoder GRU layer
    $\Gamma_{ue}$, $\Gamma_{re}$, $c_e$, $h_{(te)}$
    \State Decode to predict $Lang_B$ (English) using the encoder output of $c_e$, $h_{(te)}$
    \State Compute Compute $W1_{en}$ and $V_{en}$ for alignment model
    \State Compute $\Gamma_{ud}$, $\Gamma_{rd}$, $c_d$, $h_{(td)}$ for each
    prediction.
    \State Compute $d_1$
    \State Compute loss $L_1$ using the sparse softmax cross entropy with logits
    loss
    \State \textbf{Block 2:} Encode $Lang_B$ example and compute encoder GRU layer $\Gamma_{ue}$,
    $\Gamma_{re}$, $c_e$, $h_{(te)}$
    \State Decode to predict $Lang_A$ (Spanish) using the encoder output of
    $c_e$, $h_{(te)}$
    \State Compute $W1_{sp}$ and $V_{sp}$ for alignment model
    \State Compute $\Gamma_{ud}$, $\Gamma_{rd}$, $c_d$, $h_{(td)}$ for each
    prediction.
    \State Compute $d_2$
    \State Compute loss $L_2$ using the sparse softmax cross entropy with logits
    loss
    \State Compute total loss $L = L_1 + L_2$
    \State Optimize using Adam optimization with learning rate $\alpha$ and decayed learning rate $\gamma$.
    \EndFor
    \Until {All Phrases $Lang_A, Lang_B$ have been processed}
    \end{algorithmic}
\end{algorithm}
\FloatBarrier

This process is repeated for all the examples. Here, we keep the encoder and decoder same for all the languages that are trained for prediction. If another translation is added, then the blocks are repeated for each language.

\section{Dataset} \label{section_dataset}

Parallel datasets for Spanish and English are used for training of the Universal Vector model.  Data is taken from Many Things, an online resource for English as a Second Language Students\footnote{The primary source of the dataset used in this study as well as many more language pairings can be found at \url{http://www.manythings.org/anki/}. We used a copy hosted by the TensorFlow team at \url{http://storage.googleapis.com/download.tensorflow.org/data/spa-eng.zip} } \cite{kelly_tab_delimited_datasets_2020}. It contains 122,936 pairs of phrases in English and a corresponding Spanish translation.

\subsection{Training}

The universal vector model is trained using a modified version of the Dual Training method proposed by Xia et al. \cite{xia_2016_dual_learning}. This model is trained with a sequence for each training data, first with Spanish to English and then English to Spanish for every iteration of the dataset mentioned above. Sample phrases in both English and Spanish were used to test the predictive ability of the network into both languages. The model has been trained at 20, 30, and 40 epochs to see the effectiveness of the model as the amount of training increases. 

\section{BLEU Score} \label{section_bleu_score}

A Bilingual Evaluation Understudy (BLEU) score was used as a metric to determine the effectiveness of our NMT. BLEU was developed as a replacement for human-based validation of machine based translation that was becoming an expensive bottleneck due to the need for language expertise. The formula to calculate the score is language independent, does not need to be trained, and is able to mimic human evaluation. The function takes in the translated sentence and one or more reference sentences that it will be compared with. Groups of words, or $n$-grams, in the translated sentence to be evaluated are matched with $n$-grams in the reference sentences. 

The first step in the scoring process is to calculate a precision score by taking the number of matching $n$-grams between the evaluated sentence and the reference sentences. This number is then divided by the total count of the $n$-grams in both the references sentences and the evaluated translation. This equation can be found below in Equation~\ref{eq:bleu_precision_score} \cite{papineni_2002_bleu}. Another consideration when determining a score for a translation is the length of the output. There are many ways to say the same thing in most languages, but using too many words can introduce ambiguity and using too few words may not provide enough nuance. 

\begin{equation}
    \label{eq:bleu_precision_score}
    p_n = \frac {\sum\limits_{C \in \{Candidates\}} \sum\limits_{n-gram \in C} Count_{clip}(n-gram)}{\sum\limits_{C \in \{Candidates\}} \sum\limits_{n-gram\prime \in C\prime} Count(n-gram\prime)}
\end{equation}

Penalties are in place to ensure sentence of proper length score better. The precision score equation has a built in penalty for candidate sentences that are too long as more $n$-grams will increase the denominator and lead to a smaller score. For translations that are too short, a penalty is introduced in the form of a Brevity Penalty (BP) as in Equation~\ref{eq:bleu_brevity_penalty} \cite{papineni_2002_bleu} below. $r$ is the count of the words in the reference sentence that is closest to the translated sentence being evaluated. $c$ is the length of the candidate sentence. If there is a match, the BP is 1 and there is no penalty assessed. If there is not an exact match in length, then a penalty is assessed according to an exponentiation of $e$. 

\begin{equation}
    \label{eq:bleu_brevity_penalty}
    BP = 
    \begin{cases}
        1, & \text{if } c > r \\
        e^{(1 - r/c)} & \text{if } c \leq r
    \end{cases}
\end{equation}

The overall BLEU score for a candidate sentence is the product of the brevity penalty and the exponential sum of the product of the log of the precision score multiplied by a positive weight $w_n$. This weight is based on the number of $n$-grams $N$ such that $w_n = 1/N$. The overall score is found by using Equation~\ref{eq:bleu_overall} below. Equation~\ref{eq:bleu_overall_ranking} below is a form of the equation that is used to provide values that are more able to be ranked among other candidate translated sentences by applying a log to the whole sentence.

\begin{equation}
    \label{eq:bleu_overall}
    BLEU = BP \cdot \exp(\sum \limits_{n=1}^{N} w_n \log p_n)
\end{equation}

\begin{equation}
    \label{eq:bleu_overall_ranking}
    \log BLEU = \min (1 - \frac {r} {c}, 0) + \sum \limits_{n=1}^{N} w_n \log p_n
\end{equation}

The NLTK BLUE Score package is used for evaluation of the model\footnote {documentation for the BLEU score functionality can be found \url{https://www.nltk.org/\_modules/nltk/translate/bleu\_score.html}}. 

\section{Results} \label{section_results}

The following section will cover the translation results obtained from the Universal Vector model. It will discuss the translations from English to Spanish and Spanish to English. 

\subsection{Translations}

Example phrases in each language were fed to the model. Two example pairs of phrases are found in Table~\ref{tab:example_sentences} below.

\begin{table}[!htbp]
\centering
\caption{Example phrases used for testing}
\label{tab:example_sentences}
\begin{tabular}{|l|l|}
\hline
\textbf{English}  & \textbf{Spanish}\\
\specialrule{.2em}{.1em}{.1em}  
They abandoned their country & Ellos abandonaron su país\\
\hline
This is my life & Esta es mi vida\\
\specialrule{.2em}{.1em}{.1em} 
\end{tabular}
\end{table}

The results of the English to Spanish task can be found in Table~\ref{tab:table_english_to_spanish_results} below. In the case of our model, a BLEU score cannot capture the accuracy since it is based on matching $n$-grams. The sentences were too short to have anything larger than matching bigrams which are too small for the scoring algorithm. The result of the first phrase perfectly matched the reference sentence found in Table~\ref{tab:example_sentences}. The output of the second phrase switched the gender of the word for "this" in English from "esto" to "esta". Without more context before a phrase, the model is not able to consistently determine the genders of specific words.

\begin{table}[!htbp]
\centering
\caption{English input and Spanish output}
\label{tab:table_english_to_spanish_results}
\begin{tabular}{|l|l|}
\hline
\textbf{English Input} & \textbf{Spanish Output}\\
\specialrule{.2em}{.1em}{.1em} 
They abandoned their country & Ellos abandonaron su país\\
\hline
This is my life & Esto es mi vida\\
\specialrule{.2em}{.1em}{.1em} 
\end{tabular}

\end{table}

When English and Spanish are flipped, the model provided similar results. The resulting English outputs can be found below in Table~\ref{tab:table_spanish_to_english_results}. Small differences are present, again due to small gender differences that small sentences will be expected to yield without proper context for pronouns. 

\begin{table}[!htbp]
\centering
\caption{Spanish input and English output}
\label{tab:table_spanish_to_english_results}
\begin{tabular}{|l|l|}
\hline
\textbf{Spanish Input} & \textbf{English Output}\\
\specialrule{.2em}{.1em}{.1em} 
Ellos abandonaron su país & They abandoned his country\\
\hline
Esta es mi vida & This is my life\\
\specialrule{.2em}{.1em}{.1em} 
\end{tabular}
\end{table}

Sentences longer than four or five words yielded very poor results. This is due to the small dataset and low number of training iterations when compared with other papers in the NMT space such as most of those cited in this paper. With a larger dataset and more training time the model will better handle longer phrases.

\section{Model Analysis} \label{section_model_analysis}

The following section covers the analysis and each subsection that follows is the analysis discussion for the BLEU score, Loss analysis and Attention model.

\subsection{BLEU Score Analysis}

Applying the BLEU score to the Universal Vector Model resulted in unfavorable scores. Table~\ref{tab:BLEU} shows the results of the BLEU score from Spanish to English and English to Spanish. BLEU score calculations are provided as part of this work to show the minimum capability of this model to translate more than one language using a single universal model. The score from this work should not be compared with other translation models like Bert and other Transformer based models \cite{Vaswani_2017_attention_is_all_you_need,zhu_incorporating_bert_2020}. There are two main reasons behind this. First, the  tested  sentence  words  were  short  in  length. Second, the  short  sentences did  not  meet  the  minimal  n-gram  length  of  2  for  proper  scoring. The  use  of longer sentences could have solved these issues, however the model had difficulty translating longer sentences at the level of training we were able to accomplish in the time given (60 epochs).

\begin{table}[!htbp]
\small
\centering
\caption{BLEU Score Results}
\label{tab:BLEU}
\begin
{tabular}{p{3.5cm}|p{3.5cm}|p{3.5cm}}
\hline
\specialrule{.2em}{.1em}{.1em} 
\textbf{Sentence} & \textbf{Direction}  & \textbf{BLEU Score}\\
\specialrule{.2em}{.1em}{.1em} 
esto es mi vida. & English to Spanish  & \textbf{8.3882e-155}\\
\specialrule{.2em}{.1em}{.1em} 
this is my life. & Spanish to English  & \textbf{6.8681e-78}\\
\specialrule{.2em}{.1em}{.1em} 
\end{tabular}
\end{table}
\FloatBarrier

\pagebreak

\subsection{Loss Analysis}

\begin{figure}[!htbp]
    \includegraphics[width=10cm]{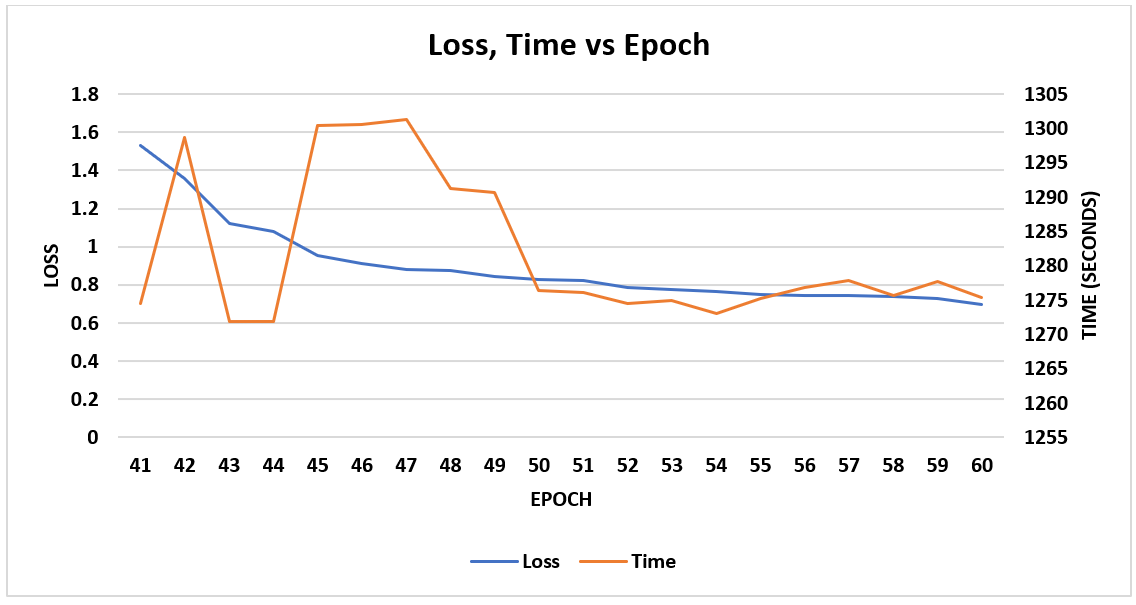}
    \centering
    \caption{Loss analysis by epoch during training of the model.} 
    \label{loss_time_epoch}
\end{figure}
\FloatBarrier

Since the BLEU score could not properly capture model accuracy for testing, more attention has been placed into minimizing loss. The loss explains how well the model is performing by minimizing error. A lower number for loss correlates to a better performing model. Figure~\ref{loss_time_epoch} shows the performance of the universal vector model by loss and training time over each epoch. The results of the model are shown between 40-60 Epochs to show where the loss curve flattens. The figure shows that the loss gradually declined between 41 to 48 epochs and began to stabilize at about $.6-.7$. 

At the 60th epoch, a loss of 0.6963 was obtained which was sufficient to translate short sentences. The model struggled with training performance with respect to time between 40-50 epochs. The exponential jump in time could potentially be due to the model struggling to get to the local minima point during optimization. Overall the loss obtained is sufficient to translate short sentences and shows that the universal vector model can translate words with minimal error.

\subsection{Attention Model Analysis}

\begin{figure}[!htbp]
    \includegraphics[width=9cm]{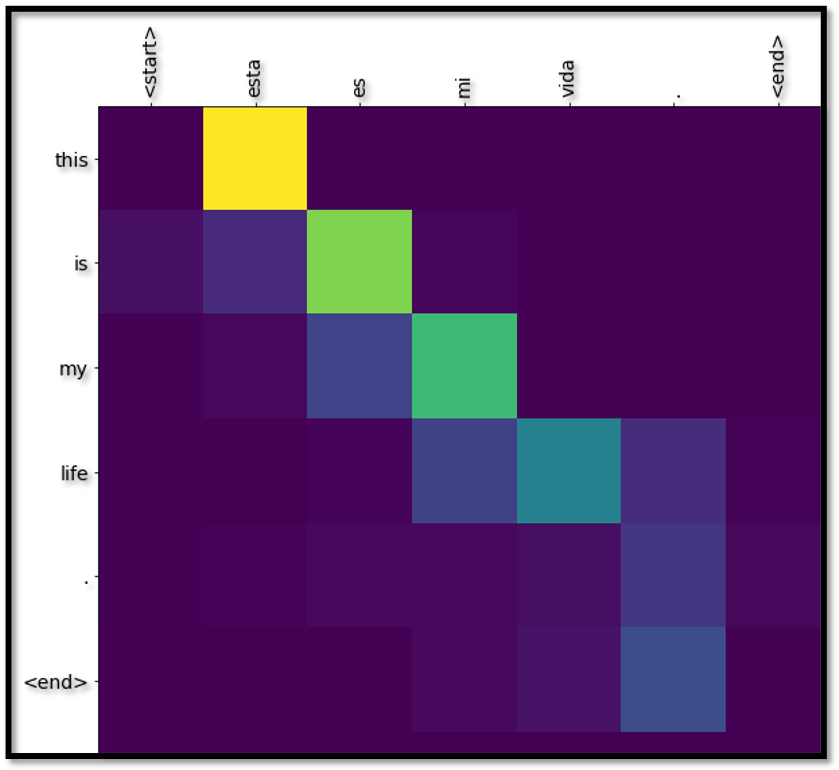}
    \centering
    \caption{Heat map showing areas of focus from Spanish to English.} 
    \label{attention_heatmap_sp_to_eng}
\end{figure}

Heat maps were created to visualize how the attention mechanism directed the focus of the decoder when predicting the corresponding text in a translation. The diagram has each word in the source language on the top and each word in the predicted sentence in the target language on the left axis. Fig.~\ref{attention_heatmap_sp_to_eng} below was generated when the Spanish phrase "Esta es mi vida" was fed into the model. On the left is the output of the model which is a prediction of the English Translation. As a visual reminder, the heat map does not necessarily show how words are correlated from source to target. Instead the visual representation of the heat map gives insight into the parts of input that the attention model focuses on when translating. For example, the yellow box in the upper left shows heavy focus on the Spanish word "esta" when the model predicts the English word "this". From there, the heavy areas of focus follow a diagonal line down and to the left. This means as the decoder moves on to predict words later in the sentence, the focus is directed to later parts of the source sentence which is generally good. Longer sentences would show more defined and more varied areas of heat as they get more complicated. Overall, the maps generated from the small sentence sizes that the model can handle, show potential that the modified attention mechanism is working as intended.

\FloatBarrier

\section{Limitations and Future Expansion} \label{section_limitations_and_future_expansion}

For further model experimentation on translation of more than two languages, a parallel dataset containing a triad of language phrases is required. While the architecture and model as part of this experiment is created to handle more than two languages, we only consider using a single model for two languages. As of today, most parallel datasets available are bilingual. In the future, a parallel dataset with three or more languages will be used to train and modify the current universal vector representation model. Furthermore, larger datasets will be used with more training iterations akin to other papers in the NMT space. A more standardized test such as those provided by the annual Workshop on Machine Translation can be used on the model translated text. 

\section{Previously Considered Experiments} \label{section_previously_considered_experiments}

Connected Learning was the first attempt at a novel proposal. During the time of initial research, there were no other papers proposing the methods that made up this new idea. This method would allow the weights to learn source and target language as a $Z$ format. First the model is trained in the direction of Source $\rightarrow$ Target, then immediately trained again with the direction of Target $\rightarrow$ Source, and finally the weights are retrained from Source $\rightarrow$ Target. 

In connected learning, training is done on the source sequence of vectors as $x = (x_1, \dots , x_{T_x})$ and target sequence of vectors as $y = (y_1, \dots, y_{T_x})$.  For each of the sequence pairs of vectors, the source and target are swapped twice by utilizing the hidden output as the input when swapped. For example, if vector sequence $x$ represents Spanish and vector sequence $y$ is English, the model would first generate $h_{(t)} = f(h_{(t-1)}, x_t)$ and $c = q(h_1, \dots, h_{T_x})$ and use the context vector "$c$" when combining the hidden state of the recurrent network and provide vector sequence $y$ as the source and generate values for $x$.

The belief was that the weights in the contextual information would have all the target information, however the model could not converge to a local optimum point where it was aligned to both source and target information.

\section{Conclusion} \label{section_conclusion}

In this paper, the idea of a "Universal Vector" is proposed as a new facet of NMT that can be used to translate between multiple languages in the same vector space. Models are usually built to translate in one direction. There exists some work that has been done in using both directions between a source and target language for reinforcement learning of training sets. However, the "Universal Vector" model is a singular model that can be trained in both directions (source to target and target to source) for more than one pair of languages. 

The "Universal Vector" model detailed in this paper was built to test the proposition by modifying an RNN based Encoder-Decoder model. Existing attention mechanisms were also modified and used to create context vectors that increased performance in predicting the next translated text for overall target phrase translation. Multiple fully connected layers are added, one for each target language, to facilitate translations into multiple target languages. 

The model is trained with parallel English and Spanish datasets. Phrases from both languages are trained from English to Spanish and Spanish to English within a recurrent network using Dual Training based methods. It was tested with many examples of both Spanish and English phrases. The attention mechanism was evaluated by viewing heat maps of where the model selectively focused on input text for its corresponding translated text.

While the results are promising, with more time and resources the experiment would provide better results. With more computing power the model can be trained using more words with more languages in a reasonable amount of time. In the future, better accepted benchmarks in translation such as those provided by the annual Workshop on Machine Translation can be used. While limited in scope, these results point to potential for greater accuracy on using a singular model for translating between multiple languages. 

%
%
%

\bibliography{jrs_final}

\end{document}